\documentclass{article}

\usepackage[nonatbib,preprint]{neurips_2021}

\usepackage[utf8]{inputenc} %
\usepackage[T1]{fontenc}    %
\usepackage{hyperref}       %
\usepackage{url}            %
\usepackage{booktabs}       %
\usepackage{amsfonts}       %
\usepackage{nicefrac}       %
\usepackage{microtype}      %
\usepackage{xcolor}         %

\usepackage{caption}
\usepackage{subcaption}
\usepackage{amsmath,amssymb}
\usepackage{hyperref}
\usepackage[capitalise]{cleveref}
\newcommand{\citep}{\cite}
\usepackage{graphicx}
\usepackage{booktabs} 
\usepackage{adjustbox}

\usepackage{contour}
\usepackage{ulem}
\contourlength{0.8pt}
\newcommand{\myunderline}[1]{%
  \uline{\phantom{#1}}%
  \llap{\contour{white}{#1}}%
}

\title{Global explainability in aligned image modalities}

\author{%
  Justin Engelmann \\
  UKRI CDT Biomedical AI\\
  University of Edinburgh\\
  \And
  Amos Storkey \\
  School of Informatics\\
  University of Edinburgh \\
  \And
  Miguel O. Bernabeu \\
  Usher Institute \\
  University of Edinburgh \\
}

\begin{document}

\maketitle

\begin{abstract}
    Deep learning (DL) models are very effective on many computer vision problems and increasingly used in critical applications. They are also inherently black box.
    A number of methods exist to generate image-wise explanations that allow practitioners to understand and verify model predictions for a given image. 
    Beyond that, it would be desirable to validate that a DL model \textit{generally} works in a sensible way, i.e. consistent with domain knowledge and not relying on undesirable data artefacts. For this purpose, the model needs to be explained globally.
    In this work, we focus on image modalities that are naturally aligned such that each pixel position represents a similar relative position on the imaged object, as is common in medical imaging. 
    We propose the pixel-wise aggregation of image-wise explanations as a simple method to obtain label-wise and overall global explanations. 
    These can then be used for model validation, knowledge discovery, and as an efficient way to communicate qualitative conclusions drawn from inspecting image-wise explanations.
    We further propose Progressive Erasing Plus Progressive Restoration (PEPPR) as a method to quantitatively validate that these global explanations are faithful to how the model makes its predictions. We then apply these methods to ultra-widefield retinal images, a naturally aligned modality. We find that the global explanations are consistent with domain knowledge and faithfully reflect the model's workings. 
\end{abstract}

\section{Introduction}

\subsection{Motivation}

Deep learning (DL) models provide excellent performance on many computer vision problems.  Accordingly, they have risen in popularity and are increasingly used in practice. This includes critical applications like healthcare where it is very important to understand and validate DL models. Unfortunately, DL models are black boxes that are inherently hard to explain, especially compared to traditional approaches such as using linear models or small decision trees with hand-crafted image features. DL models typically have millions of parameters and complex architectures with many non-linearities \citep{choo2018visual}. Thus, there is a pressing need for explainability. A number of approaches 
for image-wise explanations have emerged that can help practitioners understand why a model made a specific prediction for a given image. This allows to validate the model's prediction and to guide an expert's attention to regions of interest. 

While image-wise explanations can potentially be of great value and even indispensable in some applications, they also have some drawbacks. For one, an image-wise explanation can only be generated once the image in question has been observed. However, at that point in time no expert might be available and examining an image-wise explanation for every prediction made might be too labour-intensive to be feasible. Thus, we would like to explain the model globally in order to validate that it behaves in a sensible way that is consistent with domain knowledge and does not show signs of leveraging undesirable data artefacts \citep{roberts2021common}, also known as ``shortcuts'' \citep{geirhos2020shortcut,degrave2021ai}.
Such a global explanation could give us confidence that the model is generally working correctly and thus lessens the need for examining image-wise explanations for each new prediction.

In this work, we introduce the notion of an aligned image modality and propose to leverage this alignment to generate label-wise and overall global explanations by aggregating image-wise explanations. These global explanations can then be qualitatively compared with domain knowledge to validate that the examined DL model works correctly. They could also be used for knowledge discovery and serve as a space-efficient way to summarise image-wise explanations that is less susceptible to cherry-picking and confirmation bias than just presenting a few examples. Finally, we propose Progressive Erasing Plus Progressive Restoration (PEPPR) as a way to quantitatively verify that these global explanations faithfully reflect how the model makes its predictions and to investigate which image regions contain information about the target variables. Similar erasure-based tests have been proposed in the context of image-wise explanations \citep{samek2016evaluating}, but to our knowledge those erase one way (most to least important) rather than both ways. Furthermore, we propose to consider label-wise metrics rather than aggregate metrics which allows to find label-specific issues.

\subsection{Explainability - What, why, and a brief taxonomy}

What constitutes an explanation is a complex philosophical issue in its own right \citep{bromberger1992we, thagard1978best} and discussing this in detail is beyond the scope of the present work. Briefly, the concept of an explanation is related to the concept of understanding. Broadly speaking the goal of explainable AI in this context is to explain how a DL model works, and thus to increase people's understanding of how the model works. Depending on who is addressed, different explanations will be suitable. However, there is an objective and a subjective criterion at play: The goal of an explanation is not merely to increase someone's understanding of the model objectively but also to increase their trust by increasing their subjective confidence in their understanding. This could include a practitioner in the application domain who is perhaps not very knowledgeable about DL, but also the DL practitioner themselves who aims to understand the DL model better to ensure that it works correctly so that they can recommend its use in application in good faith. However, while increasing trust is a legitimate aim, we only want to generate well-founded, appropriate trust. Explanations can be misleading and generate unfounded trust \citep{lakkaraju2020fool}, and this must be avoided.

There are many subtly different questions and thus different senses of explainability tangled up in this broad aim of explaining ``how a DL model works''. Each sense is tied to a different concrete purpose and calls for different methods. For the present work, we primarily distinguish two types of explainability: local and global. 
Local explainability is concerned with explaining the model's prediction for a specific example, i.e. the question of \textit{``What in this particular example does the model consider evidence for its prediction?''} As a brief note on terminology: we consider ``local'' to be the natural counterpart to ``global'' but it could also be understood to refer to a specific region of the data manifold. Thus, we try to use the term ``image-wise'' to avoid this potential confusion. For computer vision, image-wise explainability is commonly accomplished through generating so-called saliency heatmaps that highlight image regions that were key for the model's prediction. Such image-wise explanations are particularly useful if critical decisions are to be based on the model's prediction. For example, if the model is used to assist a clinician in assessing medical images and the model predicts the presence of disease in a scan. Here, a local explanation allows the clinician to try to comprehend the model's prediction and draws their attention to regions of possible pathology. 

Global explainability, on the other hand, is concerned with explaining how the model generally works, i.e. the question of  \textit{``What does the model tend to focus on when making a particular kind of prediction?''} Global explanations can help us understand the model generally which allows to validate that it works in a desirable fashion and is thus suitable for being applied in practice. For instance, if the model generally focuses on image features that contain information about the target variable, this would suggest that it works as desired; whereas if it focuses on features that should not be informative about the target variable, this would indicate that the model might be relying on undesirable shortcut artefacts. These artefacts are informative in the training data but might be uninformative or simply not present during inference, leading to severely degraded model performance \citep{geirhos2020shortcut,degrave2021ai}. 

Manually assessing many image-wise explanations can build towards global explanations. However, this approach is labour intensive and examining all images is not feasible for modern image datasets that can contain anywhere between tens of thousands and hundreds of millions of images \citep{sun2017revisiting}. Conclusions that are drawn from examining a small number of examples, however, are susceptible to biases such as confirmation bias. Real world datasets and image-wise explanations can both be noisy in their own right. This could then lead to an actual yet unexpected pattern being dismissed as being spurious, whereas an expected yet spurious pattern is accepted as an actual pattern. This motivates the need for methods for global explainability. 

The distinction between image-wise (or local) and global explainability is the most relevant to the present work. There are a few additional distinctions that are also relevant. First, the explainability methods we consider here are all post-hoc rather than ante-hoc in the sense that they explain a black box model after it has been trained. However, global explainability as we are considering it here can be used to explain the model generally and thus increase our confidence that it will work correctly before new instances are observed during inference. While image-wise explanations are also very useful in practice to validate a particular model decision, not all use-cases of DL models might have an expert present at inference time. 

Another distinction is between model-centric and data-centric approaches. We focus on validating a particular DL model in this work and thus on model-centric explanations. However, data-centric explainability is also useful as it can help validate a dataset and allow for knowledge discovery. In the next section, we briefly explain how our methods could be easily modified to be data-centric instead. Finally, an explanation can be label-wise or general. We use the term ``label'' to refer to individual values of the target vector in a general sense, as opposed to ``class'' which usually implies that only a single value of the target vector can be ``hot'' at a time.

\subsection{Aligned image modalities \& aggregated image-wise explanations}

\begin{figure}[!t]
\centering
\includegraphics[width=\textwidth]{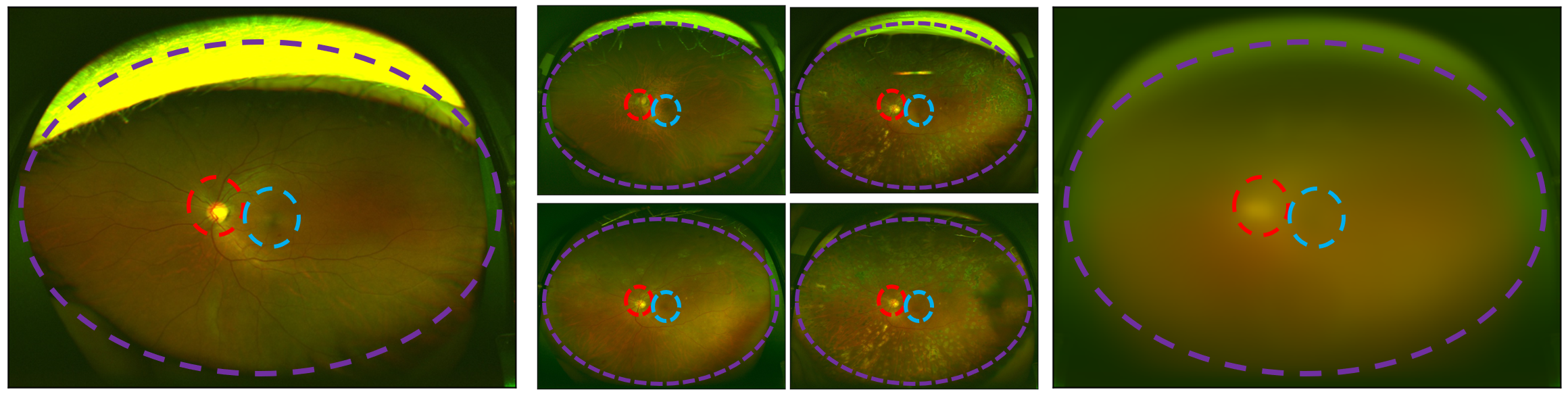}
\caption{\textbf{Example of an aligned image modality.} In an aligned image modality (here: ultra-widefield retina images), each pixel corresponds to a similar position of the imaged object (i.e. human retinas). \textbf{\protect\myunderline{Left:}} An example ultra-widefield retina image with the approximate locations of three landmarks indicated by dashed lines: \textcolor{purple}{The retina itself}, outside of this region no relevant information should be found; \textcolor{red}{the optic disc}, where blood vessels go through the retina, visible as a bright spot; \textcolor{cyan}{the fovea}, a small pit responsible for the sharpest vision, visible as a dark spot. \textbf{\protect\myunderline{Middle:}} Four more images with the same landmark indicators at identical coordinates. \textbf{\protect\myunderline{Right:}} The pixel-wise average of 1,714 validation set images. We can make out the same landmarks here (even the fovea, faintly, when zoomed in), indicating that these images are well-aligned.}
\label{fig:alignedplot}
\end{figure}

An image modality is aligned if each pixel/voxel corresponds to the same relative position of the imaged object and thus samples implicitly share a common coordinate system. Natural image datasets are rarely aligned, but fortunately such alignment is common in medical imaging where the need for model validation and potential for knowledge discovery is particularly large. For instance, in brain Magnetic Resonance Images each voxel corresponds to the same part of the brain across scans after registration. Even without explicit registration, other types of medical images tend to be naturally aligned such as chest X-rays, or can be trivially aligned.
\cref{fig:alignedplot} illustrates this for ultra-widefield retina images. After flipping all right eyes horizontally, the images are aligned such that relevant regions of the retina appear share coordinates across images, including but not limited to the visually apparent landmarks indicated in \cref{fig:alignedplot}. Image modalities with fixed reference frames like camera data from a self-driving car might also exhibit alignment. Even modalities that are not intrinsically aligned could be aligned through pre-processing. For instance, we could first use a model to detect objects, crop them, and then input these objects into an object identification model, which might then deal with aligned inputs.

We propose to leverage such alignment by pixel-wise aggregating image-wise explanations into global explanations. As the same pixel position corresponds to the same relative position of the imaged object, these aggregated image-wise explanations should then highlight which regions of the input images are generally used by the model to make its predictions. For a model that performs better than random chance, this is then a measure of where in the input images information about the target variables is found that is also used by the examined model. In principle, we could also extend this approach by then aggregating model-wise global explanations across different models to move from model-centric to a data-centric global explanations. In the present work, we focus on the model-centric explanations.

\subsection{Related work}

A large literature on explainable AI has emerged \citep{gilpin2018explaining}, with many methods focusing on post-hoc explanations of black box models \citep{guidotti2018survey}. Global explanations have been considered before in the context of tree-based algorithms applied to tabular data \citep{lundberg2019explainable}. For DL models in computer vision, a number of methods for image-wise explanations have emerged (e.g. \citep{shrikumar2017learning, selvaraju2017grad, lundberg2017unified}). Some data-centric methods go in the direction of global explanations, such as learning a generative model or a cycle-consistent image-to-image translator to identify key distinguishing features of different datasets (e.g. \citep{degrave2021ai}). In terms of explaining a fitted DL model generally, feature map visualisations \citep{olah2017feature} and image generation from classifiers \citep{pal2021synthesize} are two kinds of approaches towards global explainability. However, applying them to validate a model has some challenges as there are many feature maps that can be visualised and a whole space of class conditional samples that can be generated. Our approach generates a single global explanation per label and a single overall global explanation. Furthermore, our approach generates a spatial explanation. In aligned image modalities, different areas of the images contain different information and specific labels might occur in different regions. Our methodology allows to investigate this spatial dependence. 

\section{Methods}

\subsection{Image-wise explanations}
For a given fitted DL model $f_\theta$ with model parameters $\theta$, an image- and label-wise explanation method takes an input image $X_i$ and a target label $l$ as inputs and yields an explanation $E^l_i$ for the model's predictions of the target label for this image $\hat{p}_i(l)=(f_\theta(X_i))_l$. The explanation $E^l_i$ is a matrix of pixel-wise importance scores with the same dimensions as the input image where higher values indicate that an input pixel was more important to the model's prediction of the target label. 
We use Gradient-weighted Class Activation Mapping (GradCAM) \citep{selvaraju2017grad} to generate image- and label-wise explanations because it is a well-established method (e.g. \cite{matsuba2019accuracy, nagasato2019deep, nagasato2018deep}) and because it is generally faithful to how the explained model works. Other methods have been shown to be akin to simple edge detectors and to yield very similar heatmaps even when replacing the trained model weights with random weights \citep{adebayo2018local,adebayo2018sanity}. Human vision is generally biased towards edges and thus merely highlighting edges on an image might appear to be a reasonable explanation of the model's prediction, even if it is not actually faithful to the workings of the model. This also motivates the need for validating the obtained heatmaps beyond manually assessing a few examples.

\subsection{Global explanations}
\label{sec:methods_global}
\begin{figure}[!t]
\centering
\includegraphics[width=\textwidth]{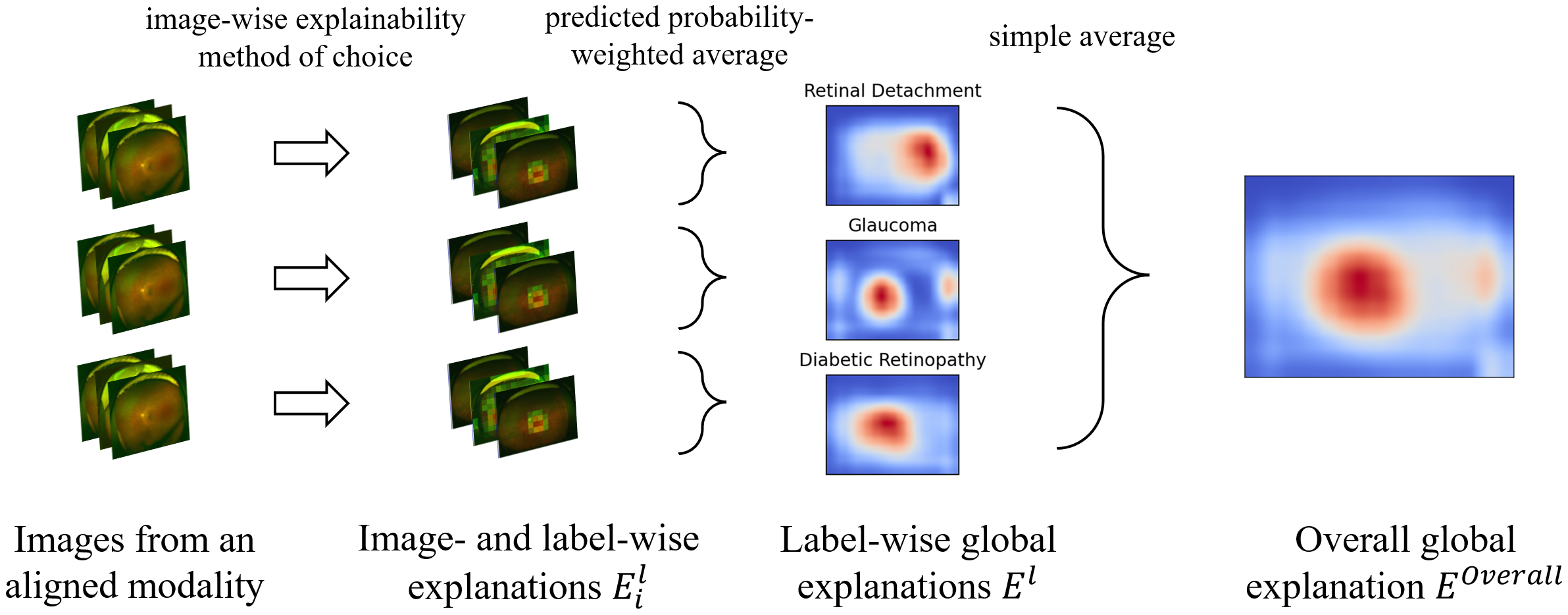}
\caption{\textbf{From image-wise to global explainability through aggregation.} In an aligned image modality, the pixel-wise aggregates of image-wise explanations can be used as global explanations.}

\label{fig:ideaplot}
\end{figure}

We aggregate image- and label-wise explanations into a label-wise global explanation. Specifically, we take the average of all true positive instances from a validation or test set weighted by the model's predicted probability of the target label $\hat{p}_i(l)$. We denote the number of positives instances of the target label as $n(l)$.

\[E^l=\frac{1}{n(l)} \sum_i^{n(l)} \hat{p}_i(l) \kern0.2em E^l_i\]

We weigh by the predicted probability $\hat{p}_i(l)$ as image-wise explanations for examples where the model does not predict the target label with high confidence are very noisy. This is expected as such an image- and label-wise explanation is also conceptually unsound if the model does not think that the target label applies. Thus, we assign less weight to them. We further only include positive instances of the target label, as otherwise the global explanations for rare labels could be dominated by noisy image-wise explanations where the model does not predict the target label with high confidence. Predicted probability weighting does mitigate this but not fully, especially if the DL model is not well-calibrated. Using a validation or test set instead of training examples avoids aggregating spurious patterns of a model that has started to overfit. Generating global explanations using the training set and comparing them to those obtained on the validation or test set could help diagnose what a model tends to overfit on but we will leave this for future work.

These label-wise global explanations $E^l$ can then be further aggregated into an overall global explanation $E^\mathrm{Overall}$ which shows which image regions are most important to the model. We take the simple, unweighted average of label-wise global explanations to preserve information about all labels even if the data is imbalanced.

\[E^\mathrm{Overall} = \frac{1}{n(\mathrm{labels})} \sum_l^{n(\mathrm{labels})} E^l\]

\subsection{Validating global explanations through Progressive Erasing Plus Progressive Restoration (PEPPR)}

The global explanations should not only appear consistent with domain knowledge, but also reflect faithfully how our model makes its predictions. Qualitative evaluation alone can be subject to confirmation biases where explanations are accepted that do not reflect the model's actual workings \citep{adebayo2018local}. To validate the global explanations quantitatively, we propose Progressive Erasing Plus Progressive Restoration (PEPPR). First, we threshold the overall global explanation at different quantiles to obtain a series of binary masks. We then use these masks to progressively erase the least important image regions globally until we are left with a blank image, and evaluate model performance at every step - without retraining as our goal is to explain and validate the fitted model $f_\theta$. Then we take the inverse of these masks, starting with a blank image, and progressively restore the least important regions. This yields two curves of threshold quantile versus performance that allow us to validate whether the global explanation is faithful to the model's workings and to better understand which image regions are informative. A detailed example in presented in \cref{sec:PEPPR_results}. We suggest erasing by either replacing the removed pixels by their average across the training set, or by random noise if the model was trained with RandomErasing \citep{zhong2020random} as augmentation. Erasure-based tests have been used to validate image-wise explanations \citep{samek2016evaluating}, but to our knowledge only starting with the most important regions and moving towards less important directions. We propose doing it in both directions so that PEPPR is sensitive to duplicated information. For instance, one area of an image might be the most important in the sense that it contains the most information about the target variable, however, some of this information might be duplicated in other image regions.

\section{Experimental results}

\subsection{Ultra-widefield retinal images}
\subsubsection{Data: the Tsukazaki Optos Public (TOP) dataset}
\label{sec:topdata_uwf_retinaldisease}
We use the Tsukazaki Optos Public (TOP) dataset \citep{ohsugi2017accuracy, masumoto2018deep, masumoto2018retinal, masumoto2019accuracy}, a dataset of 13,047 ultra-widefield retinal images.\footnote{We would like to thank Hiroki Masumoto and all his colleagues at Tsukazaki Hospital for releasing this dataset for research use. This is a great contribution to artificial intelligence research in ophthalmology.} The data was collected at Tsukazaki hospital in Himeji, Japan, between October 11, 2011 and September 6, 2018. The study was approved by the Ethics Committee of Tsukazaki Hospital (No. 191014) and the dataset is released for research use only, with commercial use being explicitly prohibited.

There are labels for eight retinal diseases, and from these we select the three most common ones: Diabetic Retinopathy, Glaucoma, and Retinal Detachment. This simplifies the discussion of our results considerably and allows us to include three characteristically different diseases. Diabetic Retinopathy manifests itself in numerous ways, primarily microaneurysms and hemorrhages which can occur across the retina, and neovascularization which occurs primarily around the optic disc \citep{Victor19}. Thus, pathology related to Diabetic Retinopathy occurs around the optic disc but is not confined to this area. Glaucoma, on the other hand, is a condition where the optic nerve is damaged and should thus be tightly localised around the optic disc where the optic nerve is situated.  Finally, Retinal Detachment can occur anywhere across the retina and unlike the other two diseases should not occur preferentially around the optic disc. However, Retinal Detachment does tend to occur more often in the superotemporal quadrant \citep{shunmugam2014pattern}, which is the top right quadrant of the retina given the way that images are shown here. Thus, we have one tightly localised disease with no signs of pathology expected elsewhere (Glaucoma), one disease that is localised with signs of pathology occurring across the retina (Diabetic Retinopathy), and one disease that is localised in a different place from the other two diseases, also with signs of pathology occurring across the retina (Retinal Detachment).

There are additional reasons to consider ultra-widefield images. It is a relatively new modality, meaning that optometrists or even ophthalmologists might not be familiar with it. The large field of view (200 degrees, compared to 30-60 degrees regular retina images) means that signs of pathology could be missed relatively easily due to the scale of the images. Thus, DL and especially model explanations could add significant practical value here. Furthermore, it has also been studied less compared to regular retina photography, so there is also more potential for knowledge discovery here.

Please note that while we chose this modality because we have a good understanding of this domain, the model we train is not intended for clinical application as it is presented here, nor do we intend to present concrete biomedical findings at present. Currently, we aim to test our proposed methodology.

\subsubsection{Datasplit and model training}
\label{sec:training_details}

The data was subset to exclude images showing other diseases, leaving 4,894 healthy images, 3,261 images with Diabetic Retinopathy, 2,440 with Glaucoma, and 933 with Retinal Detachment. We then split the data into train, validation and test sets containing 70, 15 and 15 \% of the data, respectively. We split the data on a patient- rather than image-level such that each patient occurs in exactly one of the three sets to avoid data leakage across sets. We frame the problem as multi-label classification as diseases can co-occur, using a binary target label per disease. 

As our DL model, we fine-tuned a simple ResNet18 \citep{he2016deep} using pre-trained weights from ImageNet \citep{deng2009imagenet} using the Adam optimizer \citep{kingma2014adam} with a learning rate $\eta=5\times10^{-5}$ and exponential decay rates  $\beta_1=0.9, \beta_2= 0.999$ and a label-wise binary crossentropy loss for 5 epochs which was sufficient to observe convergence. We also applied a small $L2$-penalty of $\lambda = 5\times10^{-5}$. The batch size was set to 32 due to memory limitations and training took about 7 minutes per run using a single NVIDIA RTX 2060 6GB. We use RandomErasing \citep{zhong2020random} as our only data augmentation, which with probability $p=0.3$ randomly replaces between 5 and 40\% of the image with random noise with an aspect ratio between 0.3 and 3.3. We use RandomErasing because beyond the general benefits of augmentation, it also makes the model robust to the erasure parts of the input image, which might be beneficial for PEPPR. However, we do not use additional augmentations because we want to create challenging conditions for our methods and we expect that the image-wise GradCAM explanations will be less noisy the more augmentation we use during training. We chose ResNet as our DL architecture because it is wide-spread, efficient and performant. We briefly experimented with larger ResNet variants but they offered no significant performance benefit. We implemented our training process using the PyTorch \citep{NEURIPS2019_9015} and timm \citep{rw2019timm} libraries.

\subsubsection{Global explanations}
\begin{figure}[!t]
\centering
\includegraphics[width=0.85\textwidth]{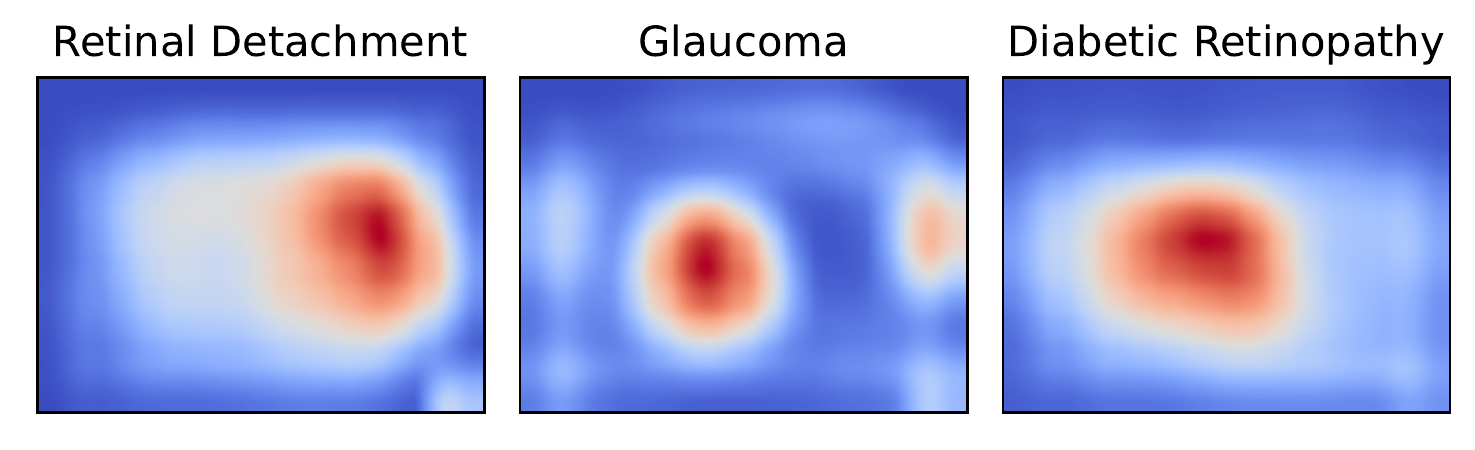}
\caption{\textbf{Label-wise global explanations.} These explanations generally match domain knowledge.
}
\label{fig:TOPmeanlabelwiseheatmaps}
\end{figure}

\begin{figure}[!t]
\centering
\includegraphics[width=0.85\textwidth]{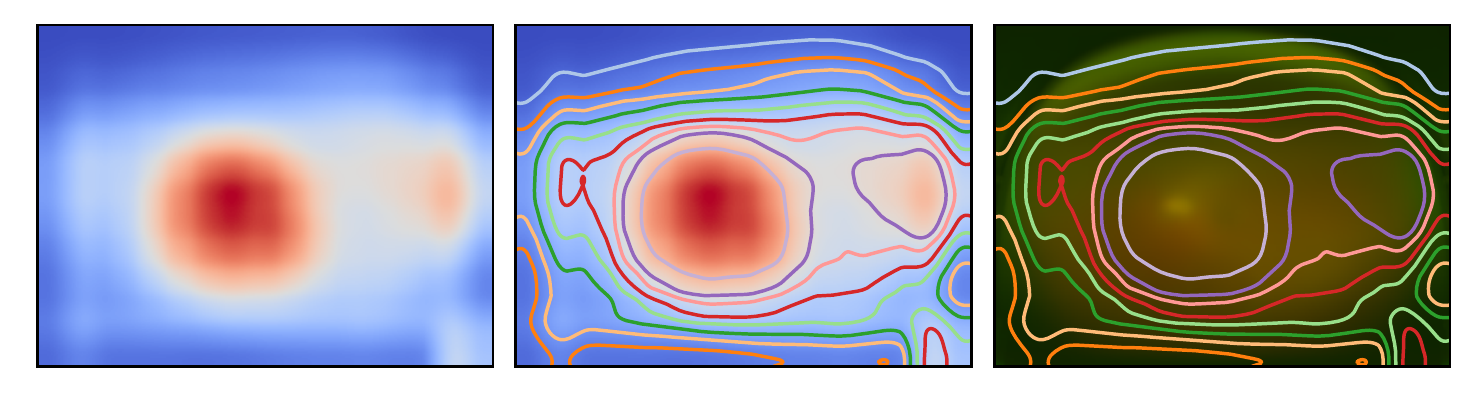}
\caption{\textbf{Overall global explanation.} \textbf{\protect\myunderline{Left:}} The overall global explanation. \textbf{\protect\myunderline{Middle:}} The overall global explanation with contour lines indicating the most important regions in quantile steps of 10\%. \textbf{\protect\myunderline{Right:}} The same contour lines overlaid on the average validation set image.
}
\label{fig:TOPoverallheatmap}
\end{figure}

We obtain the following label-wise Areas Under the Receiver Operating Characteristic Curve (AUCs) on the held-out test set: 0.9835 Retinal Detachment; 0.9180 for Glaucoma; and 0.9313 for Diabetic Retinopathy. The performance of the DL model is not the focus of this work but these values represent very good model performance and thus indicate that our model training strategy was effective. Furthermore, as the model fits the data well, the obtained explanations should reflect the relationship between data and target labels well. We generate image- and label-wise explanations for the true cases of each label as outlined in \cref{sec:methods_global} and then aggregate them into label-wise global explanations which are shown in \cref{fig:TOPmeanlabelwiseheatmaps}. We find that these generally match the domain knowledge we outlined in \cref{sec:topdata_uwf_retinaldisease}. Glaucoma is concentrated around the optic disc and has little importance allocated to other regions. Diabetic Retinopathy, too, is concentrated around the optic disc but with more importance elsewhere on the retina. Finally, the explanation for Retinal Detachment is focused on the superotemporal quadrant (upper right part of the retina) with some importance spread across the entire retina. However, these explanations also show patterns that do not match clinical evidence and thus might be signs of noise or artefacts: The explanation for Glaucoma has importance allocated to the edges of the retina, particularly to the left and right; and the explanation for Retinal Detachment has importance allocated to the bottom right corner of the images, which is an area that does not show the retina at all and thus should be uninformative. In the present work, we will use PEPPR to investigate whether the model relies on these regions. But if we wanted to apply this model in practice, then these unexpected patterns should also be investigated in more detail, for example by selecting examples where the image-wise explanations have the most importance allocated to this region. This might then reveal data issues, or potentially yield new domain insights.\footnote{For example, in the case of Glaucoma, changes on the temporal (as shown here: right) side \citep{sihota2015temporal} of the retina have been noted. We had been unaware of this previously before examining the label-wise global explanation generated with our methodology. However, we are unsure at present whether the label-wise for Glaucoma matches indeed matches this piece of clinical evidence, given that the Optical Coherence Tomography used \cite{sihota2015temporal} might have a lower field of view than these ultra-widefield retinal images. }

We also aggregate these label-wise global explanations into an overall global explanation, as shown in \cref{fig:TOPoverallheatmap}. Despite some unexpected patterns for the label-wise global explanations, the overall global explanation also matches our domain knowledge. It correctly ranks the retina regions of the image as generally more important than the non-retina regions. It identifies the area around the optic disc as the most important region, with a part of the superotemporal quadrant also ranking highly. This matches the included retinal diseases. 

\subsubsection{PEPPR}
\label{sec:PEPPR_results}
\begin{figure}[!t]
\centering
\includegraphics[width=1\textwidth]{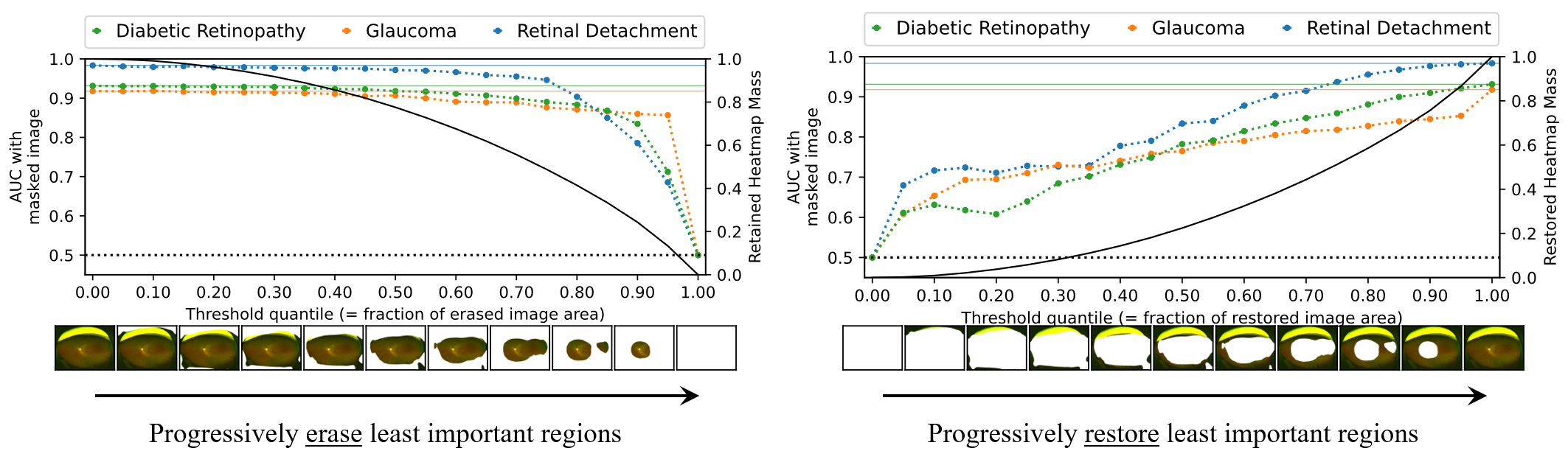}
\caption{\textbf{Results of PEPPR.} \textbf{\protect\myunderline{Both:}} Left y-axis indicates AUC obtained with the masked image. Faint horizontal lines indicate the AUC obtained with the full image. The black dotted horizontal line indicates AUC=0.5 (equivalent to random guessing). The right y-axis and the thin black line indicate the fraction of the retained importance of the global overall explanation heatmap. 
\textbf{\protect\myunderline{Left:}} Progressive erasure of the least important regions. We observe that performance for all three labels barely dropped by the time half the image was erased. For Glaucoma in particular, even only 5\% of the image are sufficient for an AUC > 0.8. \textbf{\protect\myunderline{Right:}} Progressive restoration of the least important regions. We observe a near-monotonic increase in AUC. Note the particularly large increase in AUC for Glaucoma when restoring the most important 5\% of the image containing the optic disc.
}
\label{fig:TOPPEPPR}
\end{figure}

We conducted PEPPR using quantile steps of 0.05. We replace erased pixels with random noise rather than the mean value across the training data as we trained our model with RandomErasing and thus it should be robust to encountering random noise. The results are shown in \cref{fig:TOPPEPPR}. First, we note that progressive erasure shows that the overall global heatmap does indeed faithfully reflect the model's workings. We can remove half of the image with only very small performance losses. For Glaucoma, even 5\% of the image is sufficient for an AUC>0.8, reflecting its tightly localised nature. Second, progressive restoration shows a near monotonic increase for all three labels. This suggests that while the centre region of the image contains sufficient information for high performance, some of that information might be duplicated in the periphery. Good performance on Glaucoma can be achieved with just the periphery, but performance still increases sharply when the final most important 5\% of the images is restored. However, we also note that with the least important 5\% of the images, AUCs of 0.6-0.7 can be achieved. This is unexpected as these regions do not show the retina and thus should be uninformative. This could be a sign of a data artefact that should be investigated before the model is deployed in practice. However, the progressive erasure results suggest that even if there is a data artefact in those regions, our model works well even if those regions are erased.

\section{Discussion \& conclusion}

We introduced the notion of an aligned image modality, and proposed aggregating image-wise explanations into global explanations in such modalities. We further proposed PEPPR for quantitatively validating these explanations. We then applied these methods to ultra-widefield retina images, finding that the generated global explanations are consistent with domain knowledge and that they are faithful to how the model makes its predictions. These methods are only applicable in image modalities that are aligned. This is rare for unprocessed natural images but common in domains like medical imaging or natural images that have been post-processed. Furthermore, our methods also assume that information about the target variables has a spatial dependence, 
e.g. that different labels tend to occur in different regions or that some regions should be entirely uninformative. However, in image modalities that are aligned, we would expect to find such characteristics.

While the results from our experiments are encouraging, there are many directions that future work could explore. First, we only used GradCAM to generate the image- and label-wise explanations we used as input to our methods. In principle, other methods to yield such explanations could also be used. This could yield global explanations that are more informative or similarly informative yet characteristically different to those obtained with GradCAM. Second, future work could also compare global explanations from aggregated image-wise explanations to global explanations that were generated as such. For instance, occlusion-based explainability could be directly applied on a global scale. When taking a data-centric perspective, we could frame global explainability in aligned modalities as a feature selection problem, where we want to select the most informative pixels. This might allow leveraging concepts like Relevance Determination and Max Information Gain. Third, future work could explore applying the methods to more datasets from different domains, including three-dimensional data such as brain Magnetic Resonance Images. 
Finally, there are a number of directions in which the methodology itself could be extended. For example, we could consider different aggregation functions like the median, or quantiles. Aggregation loses information and thus also aggregating by taking the pixel-wise variance or the interquartile range might yield additional insight. For instance, this could allow us to distinguish between the case where all image-wise explanations have importance evenly distributed across a region, and the case where each image-wise explanation has its importance concentrated in a different spot in that region. Using just the mean, the global explanations of the two cases could be quite similar. 

In this work, we focused on presenting and testing the methodology we introduced. We hope that these methods will be a further tool in the toolbox for model explanation, and useful to applied work where it is used to validate DL models that are developed for critical applications and  to discover new domain knowledge.

\begin{ack}
We thank Hiroki Masumoto, his colleagues Daisuke Nagasato, Shunsuke Nakakura, Masahiro Kameoka, Hitoshi Tabuchi, Ryota Aoki, Takahiro Sogawa, Shinji Matsuba, Hirotaka Tanabe, Toshihiko Nagasawa, Yuki Yoshizumi, Tomoaki Sonobe, Tomofusa Yamauchi, and the entire staff at Tsukazaki Hospital for creating the Tsukazaki Optos Public Project dataset and sharing it with the scientific community. We consider this to be a great contribution to artificial intelligence research in ophthalmology for which we are most grateful.
Funding from the UKRI CDT in Biomedical AI is gratefully acknowledged.

\end{ack}

{
\small
\bibliography{references}
}

\appendix

\section{Appendix}

\subsection{Mean retina when not flipping right eyes horizontally}
\begin{figure}[!t]
\centering
\includegraphics[width=0.5\textwidth]{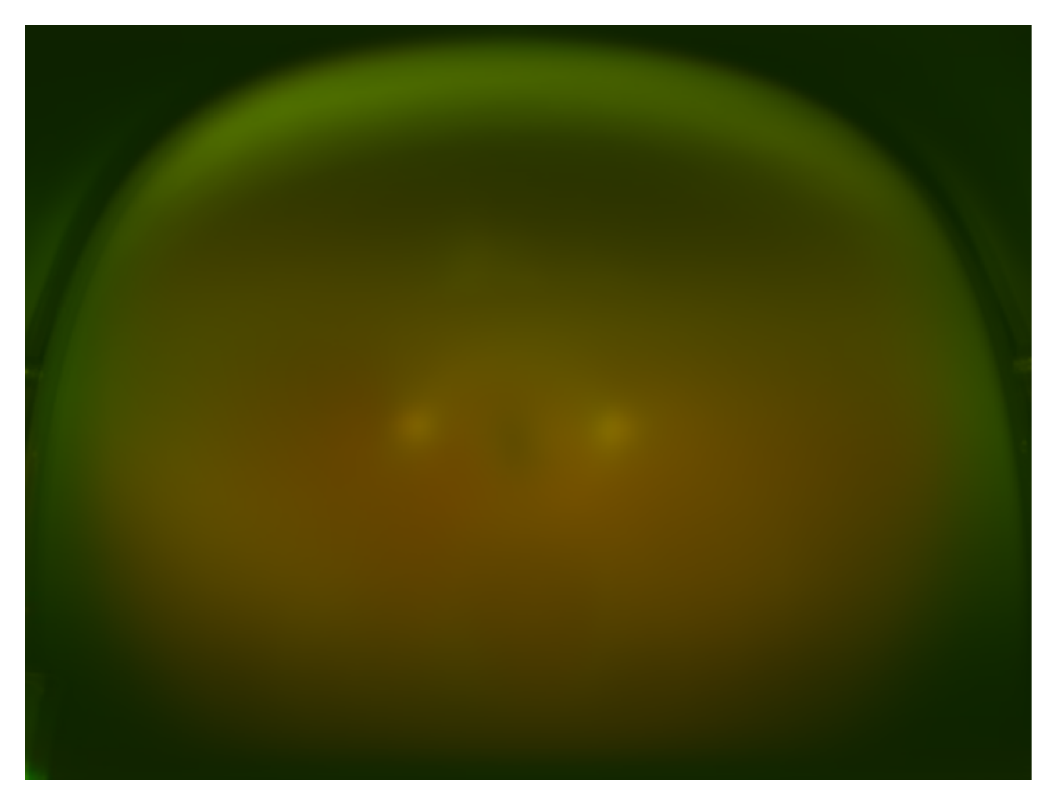}
\caption{\textbf{Mean validation image when not flipping right eyes horizontally.} The fovea (dark pit in the centre) becomes slightly easier to see, however, the optic disc (bright spot) is now duplicated. These images are less well aligned than those we all eyes were flipped to be left eyes, which is what we would expect.}
\label{fig:noleftrightflip}
\end{figure}

\end{document}